\documentclass{svproc}
\usepackage{url}

\usepackage{graphicx}
\usepackage{amsmath}
\usepackage{float}
\usepackage{dblfloatfix}
\usepackage{hyperref}
\begin{document}
\mainmatter              
\title{EViDS: An Environment--Vehicle Dynamics Integrated Simulation Platform for Intelligent Chassis Development}

\titlerunning{EViDS for Intelligent Chassis Development}

\author{
Lu Xiong\inst{1}
\and
Yiming Zhou\inst{1}
\and
Xuge Hu\inst{1}
\and
Yi Zhang\inst{1}
\and
Haoran Lv\inst{1,2}
\and
Chen~Tang\inst{1}
}

\authorrunning{Lu Xiong et al.}

\tocauthor{Lu Xiong, Yiming Zhou, Xuge Hu, Yi Zhang, Haoran Lv and Chen Tang}

\institute{
College of Automotive and Energy Engineering, Tongji University, Shanghai 200092, China\\
\email{chen\_tang@tongji.edu.cn}
\and
Shanghai Research Institute for Intelligent Autonomous Systems, Tongji University, Shanghai 200092, China
}

\maketitle            

\begin{abstract}
The development of intelligent chassis increasingly requires the integration of environmental and perception information with high-fidelity vehicle dynamics to support environment-aware intelligent chassis control. However, existing autonomous-driving simulators and professional vehicle-dynamics tools are generally developed for different purposes, making it difficult to integrate environmental information, vehicle dynamics, and chassis algorithms. This paper proposes EViDS, an Environment–Vehicle Dynamics Integrated Simulation Platform for Intelligent Chassis Development. EViDS integrates CARLA-based environment and sensor simulation with CarSim-based high-fidelity vehicle dynamics. Reference-pose-based state alignment and multi-rate interaction maintain spatial and temporal consistency, while ray-casting-based tire--road interaction transfers environmental information to the vehicle-dynamics model. Scenario configuration, synchronized data acquisition, and algorithm interfaces further support intelligent chassis development. The platform is evaluated through environment–dynamics interaction tests with vehicle-dynamics consistency comparisons, together with intelligent chassis applications. Results show that EViDS preserves vehicle-dynamics characteristics, reliably transfers environmental information, and supports online closed-loop intelligent chassis control, providing a unified simulation basis for environment-aware intelligent chassis development and validation.
\end{abstract}

\section{Introduction}

With the development of autonomous driving technologies, intelligent chassis systems increasingly require coordinated exchanges of environmental information, vehicle states, and control commands for precise motion control \cite{skrickij2024review}. Meanwhile, the growing complexity of chassis architectures and multi-actuator coordination further strengthens the need for integrating environmental information with high-fidelity vehicle dynamics \cite{lv2025adhesion}. Therefore, intelligent chassis development requires simulation frameworks that combine realistic driving environments with high-fidelity vehicle dynamics.

However, existing simulation platforms still have limitations in supporting intelligent chassis development. Autonomous-driving-oriented platforms, such as CARLA and PreScan, mainly focus on environment and perception simulation, while their built-in vehicle models provide limited fidelity for detailed chassis dynamics analysis and control validation \cite{dosovitskiy2017carla,wang2021prescan}. Vehicle-dynamics-oriented platforms, such as CarSim and Adams/Car, provide high-fidelity vehicle and tire models but generally lack realistic and configurable driving environments and perception capabilities \cite{cantas2021cosim}. Existing integrated simulation approaches still face challenges in simultaneously providing realistic autonomous-driving environments and high-fidelity vehicle dynamics \cite{nezhad2026simulators}. These limitations hinder the integrated development of environment-aware intelligent chassis algorithms.

Moreover, coupling heterogeneous simulators requires addressing differences in simulation interfaces, state representations, and time-management mechanisms to ensure consistent cross-platform interaction \cite{gomes2018cosimulation}. To address these challenges, this paper develops EViDS, an Environment--Vehicle Dynamics Integrated Simulation Platform for Intelligent Chassis Development, based on CARLA, CarSim, and MATLAB/Simulink.

The main contributions of this work are summarized as follows:

(1) An integrated environment–vehicle dynamics co-simulation architecture is proposed, enabling closed-loop interaction between autonomous-driving environments and high-fidelity vehicle dynamics for intelligent chassis development.

(2) Cross-domain interaction mechanisms are developed, including reference-pose-based state alignment, multi-rate interaction, and ray-casting-based tire--road interaction, to maintain spatial and temporal consistency and transfer environmental information to the vehicle dynamics domain.

(3) The proposed platform is evaluated through environment–dynamics interaction tests with vehicle-dynamics consistency comparisons and intelligent chassis applications, demonstrating its capability to support the development and closed-loop validation of environment-aware intelligent chassis algorithms.

EViDS will be released as open source upon publication.\footnote{\url{https://github.com/YimingZhou-tj/EViDS}}

\section{Architecture of the EViDS Platform}

\subsection{Overall Architecture}

\begin{figure*}[t]
\centering
\includegraphics[width=\textwidth]{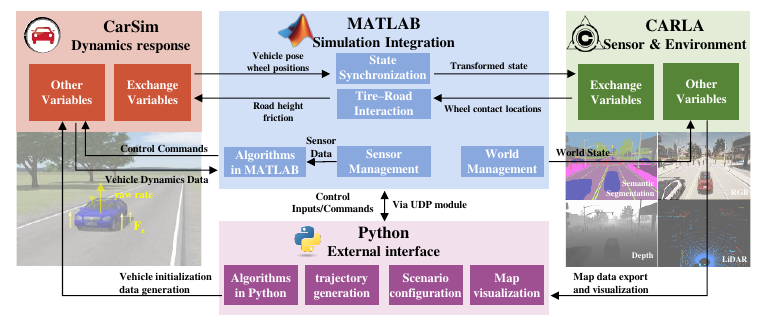}
\caption{Architecture of the EViDS platform.}
\label{fig:framework}
\end{figure*}

The proposed EViDS platform integrates high-fidelity vehicle dynamics simulation with virtual environment simulation to establish an environment--vehicle dynamics co-simulation framework, as illustrated in Fig.~\ref{fig:framework}. CarSim is connected to MATLAB/Simulink through an S-function interface for vehicle dynamics computation, while CARLA is accessed through the Python API for environment and sensor simulation. MATLAB/Simulink serves as the central integration layer for state synchronization, tire--road interaction, simulation management, and algorithm integration. The world management module handles CARLA world initialization and the creation and removal of scenario objects. Algorithms can be implemented in MATLAB/Simulink or developed externally in Python and connected through the Python interface.

EViDS data comprise cross-domain exchange variables and domain-specific internal variables. The former category includes the vehicle pose, tire-contact positions, and road information required for environment–dynamics interaction. Domain-specific variables, such as additional vehicle-dynamics states and CARLA sensor and environment data, are retained within their respective simulation domains and made available to the algorithm layer as needed. During simulation, the vehicle pose and tire-contact positions obtained from CarSim are transformed into the CARLA representation. The aligned vehicle pose is used for vehicle synchronization, while the transformed tire-contact positions are used for tire--road interaction. CARLA then returns local road height and friction coefficients to the vehicle dynamics model and provides sensor and environmental information to the algorithm layer, forming a closed-loop connection among the environment, vehicle dynamics, and chassis algorithms.

Beyond runtime co-simulation, the external Python interface provides functions for map data export and visualization, scenario configuration, trajectory generation, and vehicle initialization. The generated initialization information is transferred to CarSim to maintain consistent initial conditions between the two simulation domains. Further capabilities for intelligent chassis algorithm development are introduced in Sect.~3.

\subsection{Runtime Workflow and Key Interaction Mechanisms}

\begin{figure*}[t]
\centering
\includegraphics[width=\textwidth]{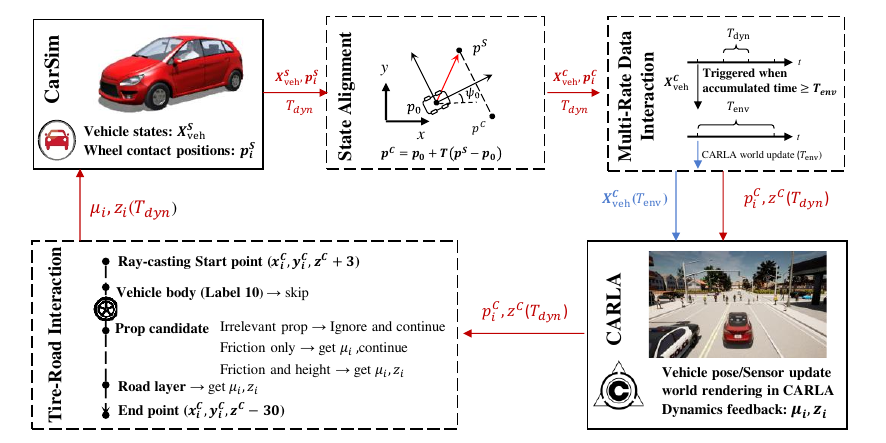}
\caption{Runtime workflow and key cross-domain interaction mechanisms of EViDS.}
\label{fig:workflow}
\end{figure*}

The runtime workflow of EViDS is illustrated in Fig.~\ref{fig:workflow}. At each vehicle dynamics step with period \(T_{\mathrm{dyn}}\), CarSim provides the vehicle pose and four tire-contact positions. The vehicle pose is denoted by
\(\mathbf{X}_{\mathrm{veh}}=[x,y,z,\psi,\theta,\phi]^{T}\),
where \(x\), \(y\), and \(z\) represent the vehicle position, and \(\psi\), \(\theta\), and \(\phi\) represent yaw, pitch, and roll, respectively. Superscripts \(S\) and \(C\) indicate the corresponding CarSim and CARLA representations. The planar tire-contact position is denoted by
\(\mathbf{p}_{i}=[x_i,y_i]^{T}\), where
\(i\in\{\mathrm{FL},\mathrm{FR},\mathrm{RL},\mathrm{RR}\}\).

\subsubsection{Reference-Pose-Based State Alignment}

Because CarSim follows the ISO coordinate convention whereas CARLA adopts the Unreal Engine convention, their pose representations cannot be directly exchanged. EViDS therefore uses the initial vehicle pose generated from the CARLA map as the common reference for state alignment and transfers it to CarSim for initialization. Let \(\mathbf{p}_{xy}=[x,y]^{T}\) denote the planar vehicle position, and let \(\mathbf{p}_{0}=[x_{0},y_{0}]^{T}\) and \(\psi_{0}\) denote the reference position and heading, respectively. The planar position mapping is expressed as
\begin{equation}
\mathbf{p}_{xy}^{C}(t)
=
\mathbf{p}_{0}
+
\mathbf{T}(\psi_{0})
\left[
\mathbf{p}_{xy}^{S}(t)
-
\mathbf{p}_{0}
\right]
\label{eq:position_mapping}
\end{equation}

The transformation matrix is defined as
\begin{equation}
\mathbf{T}(\psi_{0})
=
\mathbf{R}(\psi_{0})
\mathbf{M}
\mathbf{R}(-\psi_{0})
\label{eq:transformation_matrix}
\end{equation}
where
\begin{equation}
\mathbf{R}(\psi)
=
\begin{bmatrix}
\cos\psi & -\sin\psi \\
\sin\psi & \cos\psi
\end{bmatrix},
\qquad
\mathbf{M}
=
\begin{bmatrix}
1 & 0 \\
0 & -1
\end{bmatrix}
\label{eq:rotation_mirroring}
\end{equation}
Here, \(\mathbf{M}\) accounts for the axis-direction difference between the ISO and UE conventions, while \(\mathbf{R}\) establishes the mapping with respect to the initial heading \(\psi_{0}\).

The orientation mapping is expressed as
\begin{equation}
\left[
\psi^{C}(t),\,
\theta^{C}(t),\,
\phi^{C}(t)
\right]^{T}
=
\left[
\operatorname{wrap180}\!\left(2\psi_{0}-\psi^{S}(t)\right),\,
-\theta^{S}(t),\,
\phi_{E}^{S}(t)
\right]^{T}
\label{eq:orientation_mapping}
\end{equation}
where \(\phi_{E}^{S}\) denotes the Euler roll angle provided by CarSim for the CARLA roll update, and \(\operatorname{wrap180}(\cdot)\) limits the yaw angle to \([-180^{\circ},180^{\circ})\). The aligned pose \(\mathbf{X}_{\mathrm{veh}}^{C}\) is used for CARLA synchronization, while the transformed tire-contact positions
\(\mathbf{p}_{i}^{C}=[x_i^{C},y_i^{C}]^{T}\)
are used for tire--road interaction.

\subsubsection{Multi-Rate Data Interaction}

Vehicle dynamics computation and CARLA environment updates have different update-rate requirements. The vehicle dynamics model must be solved at a high rate to capture rapid changes in vehicle motion, whereas scene rendering, vehicle synchronization, and sensor updates can operate with a longer environment update period \(T_{\mathrm{env}}\). Tire--road information, however, directly participates in the vehicle dynamics calculation and therefore needs to be updated at the dynamics period \(T_{\mathrm{dyn}}\). Otherwise, localized variations in road geometry or surface properties may be missed between two environment updates.

EViDS therefore adopts a multi-rate interaction mechanism \cite{gomes2018cosimulation}. Simulink uses \(T_{\mathrm{dyn}}\) as the master simulation period. When the accumulated simulation time reaches \(T_{\mathrm{env}}\), the aligned vehicle pose \(\mathbf{X}_{\mathrm{veh}}^{C}\) is applied to CARLA for vehicle-pose synchronization, scene rendering, and sensor updates. In parallel, the transformed tire-contact positions \(\mathbf{p}_{i}^{C}\) are used for tire--road queries at every dynamics step, and the resulting friction coefficient \(\mu_i\) and road height \(z_i\) are returned to CarSim at the same dynamics rate. This preserves the high-rate road feedback required for vehicle dynamics computation without requiring the CARLA environment and sensors to update at the same rate.

\subsubsection{Ray-Casting-Based Tire--Road Interaction}

For tire--road interaction, a downward ray is generated at each transformed tire-contact location in CARLA. Its horizontal position is given by \((x_i^{C},y_i^{C})\), while the vehicle height \(z^{C}\) from the aligned pose is used as the reference for the vertical search range. In the current implementation, vertical offsets of \(+3\,\mathrm{m}\) and \(-30\,\mathrm{m}\) relative to \(z^{C}\) are used to define the ray start and end points, respectively. This enables local road properties to be obtained directly from the virtual environment rather than relying only on predefined road inputs \cite{sayers2011road}.

The detected intersections are processed according to their semantic labels. Intersections with the ego vehicle and irrelevant objects are skipped. For friction-only objects, \(\mu_i\) is recorded while the ray continues downward to determine the underlying road height. For objects affecting both friction and geometry, the detected friction coefficient and collision height are directly used as \(\mu_i\) and \(z_i\), respectively. When the underlying road surface is reached, its friction property and collision height are used to determine the corresponding road inputs. The resulting \(\mu_i\) and \(z_i\) are then fed back to CarSim.

Together, reference-pose-based state alignment, multi-rate data interaction, and ray-casting-based tire--road interaction establish consistent environment--vehicle dynamics interaction and enable environmental information to be incorporated into high-fidelity vehicle dynamics simulation.

\section{Functional Modules of EViDS for Intelligent Chassis Development}

\subsection{Scenario Configuration and Test Setup}

\begin{figure*}[t]
\centering
\includegraphics[width=0.98\textwidth]{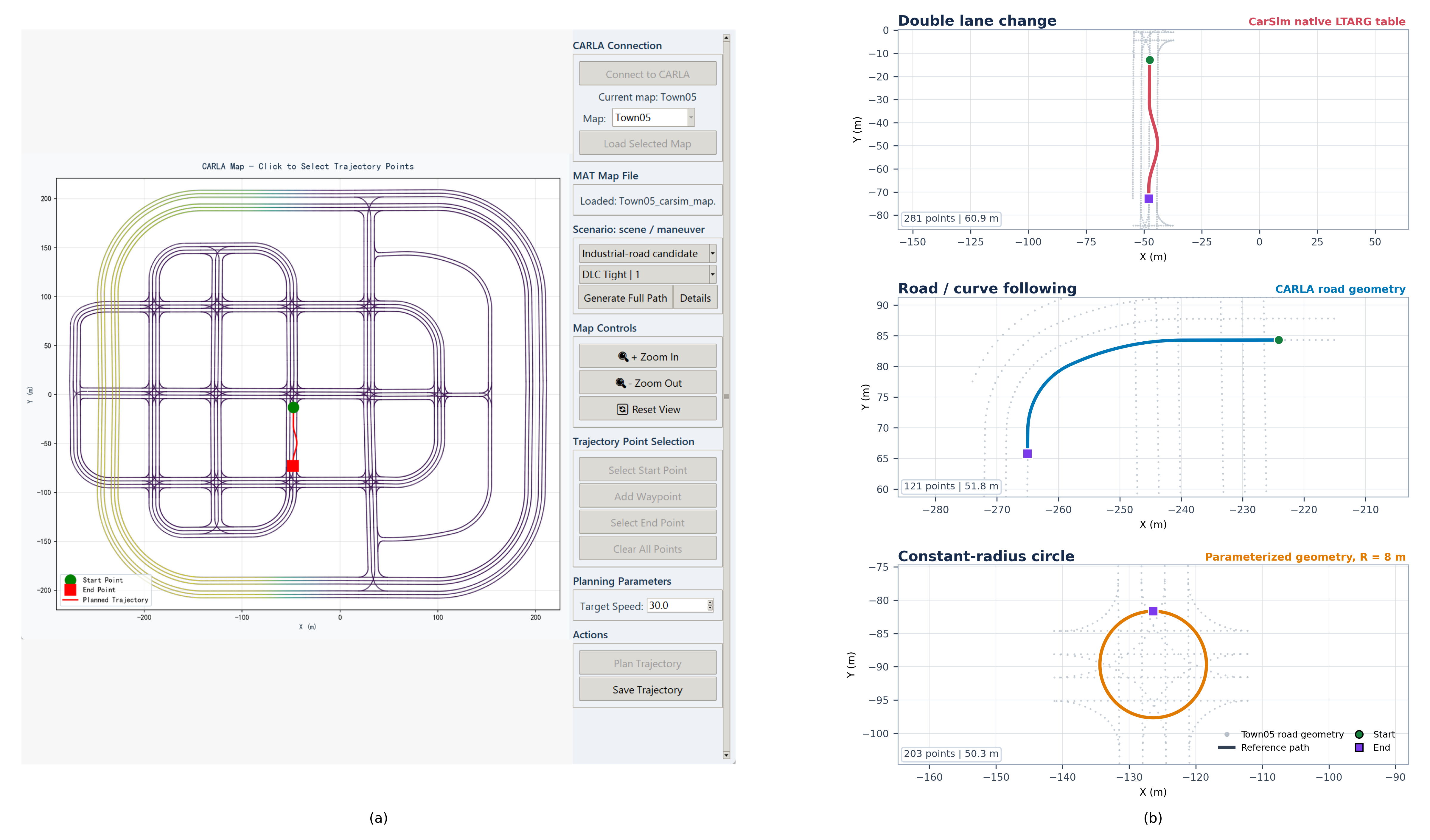}
\caption{Scenario configuration and reference paths in EViDS.
(a) Scenario configuration GUI.
(b) Representative reference paths.}
\label{fig:scenario_configuration}
\end{figure*}
EViDS provides a graphical interface for map visualization, scenario and maneuver selection, reference-path generation, and data export, as shown in Fig.~\ref{fig:scenario_configuration}(a). Conventional CarSim tests are generally conducted in simple and fixed road environments, whereas EViDS integrates the diverse virtual scenarios provided by CARLA. For environment-aware intelligent chassis algorithms, the same vehicle-dynamics maneuver may exhibit different control performance under different visual and environmental conditions. Therefore, an independent scenario library is established to manage CARLA scenarios, maneuver definitions, and their placement information, allowing different scenarios and test maneuvers to be combined into configurable test cases. The current library includes 11 scenario categories: residential, commercial, industrial, park, urban mixed, intersection-dense, highway/multi-lane, roundabout, grade/bridge, suburban/rural, and closed-test-area environments. Nine maneuver definitions are provided, including two double-lane-change configurations, single-lane change, constant-radius circular driving, lane following, curve following, grade following, roundabout following, and a legacy double-lane-change configuration. These data form 94 available map--scenario--maneuver--placement combinations.

For a selected maneuver, EViDS retrieves its placement and reference-path data, verifies consistency with the loaded CARLA map, and maps the path between the CarSim local coordinates and the CARLA world coordinates. As illustrated in Fig.~\ref{fig:scenario_configuration}(b), double-lane-change paths are generated from the original CarSim LTARG data, road- and curve-following paths are extracted from CARLA road geometry, and constant-radius circular paths are generated from parameterized geometry. The GUI also supports interactive specification of start points, waypoints, and endpoints for customized reference paths.

After configuration, the initial position and heading can be exported for CarSim initialization, while the complete reference path is retained in both CarSim local and CARLA world coordinates. This provides a consistent basis for scenario configuration, visualization, vehicle initialization, and reference-path management across the two simulation domains.

\subsection{Complex Road-Condition and Environmental Object Configuration}

Although road inputs such as elevation and friction can be predefined in CarSim, environment-aware intelligent chassis development requires not only accurate vehicle-dynamics inputs but also corresponding visual and sensor-level environmental information. CARLA provides realistic road scenes, traffic participants, and multi-source sensor simulation, making it more suitable for constructing road conditions that need to be perceived by chassis algorithms. Therefore, \mbox{EViDS} configures dynamics-related road conditions in CARLA and feeds the corresponding physical properties back to CarSim through the tire--road interaction mechanism, so that environmental perception and vehicle-dynamics inputs originate from the same virtual scene.

EViDS supports both standard and customized environment configuration. General scene objects, such as traffic cones and obstacles, can be directly deployed from the CARLA blueprint library through the Python API. For specific road-condition tests, however, existing assets may provide limited flexibility in geometry, material appearance, semantic identification, or association with predefined road properties. EViDS therefore introduces customized road-condition assets for controlled configuration of local friction variations, road-elevation disturbances, and other dynamics-related road features.

\begin{figure*}[t]
\centering
\includegraphics[width=0.98\textwidth]{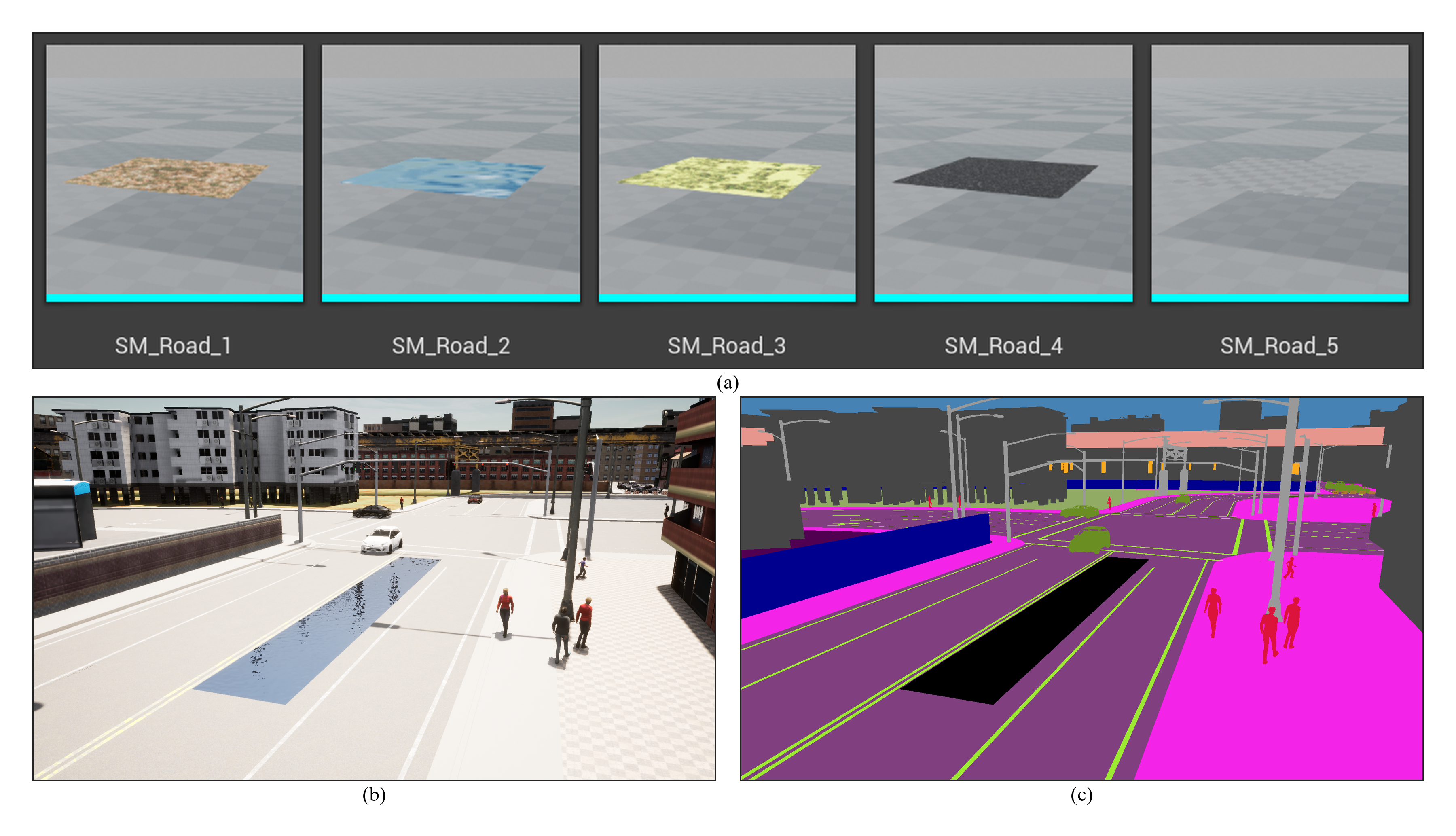}
\caption{Road-condition configuration in EViDS.
(a) Customized road-surface assets.
(b) Deployment in CARLA.
(c) Semantic identification.}
\label{fig:road_condition_configuration}
\end{figure*}

As shown in Fig.~\ref{fig:road_condition_configuration}(a), customized road-surface assets with different material appearances are integrated into CARLA and assigned corresponding semantic labels and road properties. These assets can be deployed at specified road locations according to test requirements.

Fig.~\ref{fig:road_condition_configuration}(b) shows a customized road surface deployed in CARLA, while Fig.~\ref{fig:road_condition_configuration}(c) shows its semantic representation, where the customized region appears black and is semantically distinguished from the surrounding normal road shown in purple. Based on tire-contact positions and semantic information, the tire--road interaction module identifies the configured region and obtains the corresponding friction or elevation property for vehicle dynamics calculation.

\section{Validation and Application}

The experiments were conducted using the proposed EViDS platform, integrating CARLA 0.9.11, CarSim 2019.1, and MATLAB/Simulink 2021b. EViDS inherits the environment and sensor-simulation capabilities of CARLA, allowing existing maps, traffic scenes, and environmental information to be directly incorporated into intelligent chassis development.

For environment-aware intelligent chassis development, road information that directly affects vehicle dynamics is also required. Road geometry and local friction conditions, for example, may be perceived in advance and incorporated into chassis algorithms. Therefore, this section first validates whether information about such road-condition variations can be correctly transferred from the CARLA environment to the CarSim dynamics domain, and then demonstrates the platform's support for an environment–aware perception–estimation–control application.

\subsection{Validation of Environment--Dynamics Interaction}

Two representative road conditions were considered: heterogeneous friction and local road-elevation disturbance, representing variations in road-surface properties and road geometry, respectively. In EViDS, these local road-condition variations were introduced in CARLA, detected at the corresponding tire contact positions, and transferred to CarSim as tire--road inputs. Standalone CarSim simulations with equivalent prescribed road inputs were used as references.

\begin{figure*}[t]
\centering
\includegraphics[width=0.98\textwidth]{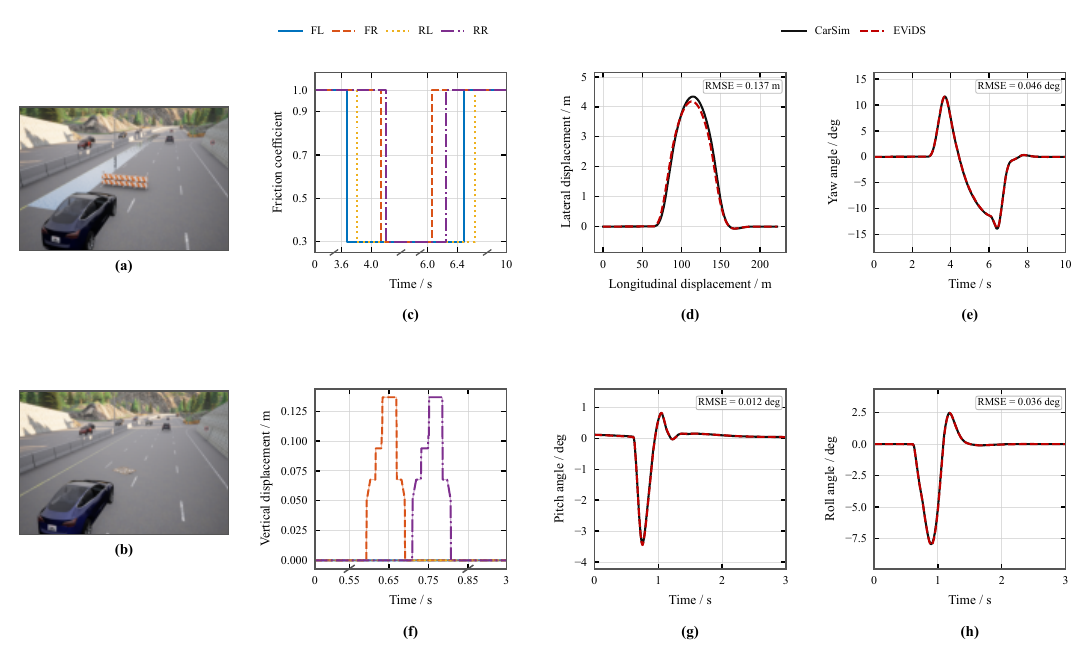}
\caption{Validation of environment--dynamics interaction.
(a) Heterogeneous-friction scene.
(b) Road-elevation scene.
(c) Tire friction feedback.
(d) Trajectory comparison.
(e) Yaw-angle comparison.
(f) Road-elevation feedback.
(g) Pitch-angle comparison.
(h) Roll-angle comparison.}
\label{fig:interaction_validation}
\end{figure*}

Figs.~\ref{fig:interaction_validation}(c)--(e) show the results under the heterogeneous-friction scenario in Fig.~\ref{fig:interaction_validation}(a), while Fig.~\ref{fig:interaction_validation}(f)--(h) correspond to the road-elevation scenario in Fig.~\ref{fig:interaction_validation}(b).

For the heterogeneous-friction scenario, the four tires enter and leave the local low-friction region at different times because of their different spatial positions. Accordingly, their friction coefficients change from $\mu=1.0$ to $\mu=0.3$ and recover after leaving the region, as shown in Fig.~\ref{fig:interaction_validation}(c). This demonstrates the spatial and temporal correspondence between the local road property and the tire contact positions.

The resulting trajectory and yaw-angle responses are compared with the standalone CarSim references in Fig.~\ref{fig:interaction_validation}(d)--(e). The two simulations show close agreement, with RMSE values of 0.137~m and 0.046~deg, respectively. This indicates that the friction information obtained from CARLA is correctly transferred to the vehicle dynamics model and produces the expected dynamic response.

For the road-elevation scenario, the local elevation disturbance is detected successively at the corresponding tire contact positions as the vehicle passes through the region, as shown in Fig.~\ref{fig:interaction_validation}(f). The resulting pitch and roll responses in Fig.~\ref{fig:interaction_validation}(g)--(h) also closely match the standalone CarSim references, with RMSE values of 0.012~deg and 0.036~deg, respectively.

These results demonstrate that EViDS can associate local road conditions with tire motion and transfer both road-surface and road-geometry information to the vehicle dynamics domain while maintaining dynamic responses consistent with the standalone CarSim references, validating the main environment--dynamics interaction mechanism described in Sect. 2.

\subsection{Intelligent Chassis Application}

\begin{figure*}[t]
\centering
\includegraphics[width=0.98\textwidth]{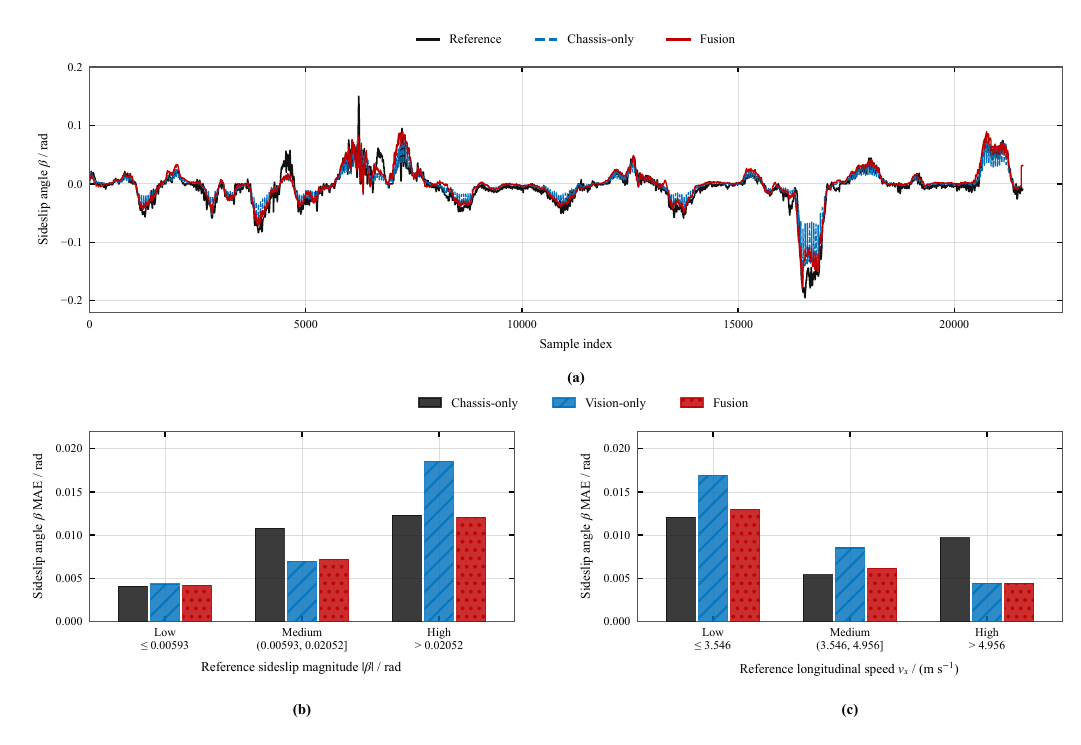}
\caption{Real-vehicle sideslip estimation under different operating conditions.
(a) Sideslip-angle estimation results.
(b) MAE under different sideslip-magnitude ranges.
(c) MAE under different longitudinal-speed ranges.}
\label{fig:intelligent_chassis_application}
\end{figure*}

To further demonstrate the applicability of EViDS to environment-aware intelligent chassis development, a perception--estimation--control application was implemented following a three-stage workflow: preliminary algorithm development and interface integration in EViDS, estimator refinement and evaluation using real-vehicle data, and online closed-loop execution after reintegration into the platform. The application serves as a platform-level demonstration rather than an independent algorithmic contribution.

Within EViDS, visual observations from CARLA and chassis signals from CarSim were used for multi-source vehicle-state estimation. Three estimation branches were considered: a chassis-only learnable Kalman filter with hybrid physics- and data-driven components, a stereo visual-odometry baseline based on RAFT-Stereo, LoFTR, and RANSAC-PnP, and a quality-aware vision--chassis multi-source Kalman fusion estimator. The chassis-only estimator combines vehicle-dynamics prediction with learnable residual correction and Kalman updates, whereas the visual branch obtains vehicle-motion states through stereo depth estimation, inter-frame matching, and pose estimation rather than directly regressing the sideslip angle from images. The fusion estimator further combines chassis observations with visual-odometry outputs through quality-aware Kalman updates.

The fused vehicle states were subsequently supplied to a motion controller with sideslip-related stability constraints, and the resulting control commands were returned to the vehicle dynamics model. This forms a complete perception--estimation--control chain through the unified data and control interfaces of \mbox{EViDS}.

Fig.~\ref{fig:intelligent_chassis_application} presents the real-vehicle sideslip-estimation results after estimator refinement. Fig.~\ref{fig:intelligent_chassis_application}(a) compares the reference sideslip angle with the chassis-only and fusion estimates, while Fig.~\ref{fig:intelligent_chassis_application}(b)--(c) compare the chassis-only, vision-only, and fusion methods under different reference sideslip-magnitude and longitudinal-speed ranges. The results show condition-dependent complementarity among the different information sources. In the highest-speed range, fusion achieves an MAE of 0.00438~rad, compared with 0.00971~rad for chassis-only estimation and 0.00442~rad for vision-only estimation. In the highest-sideslip-magnitude range, fusion also yields the lowest MAE of 0.01208~rad. The chassis-only and vision-only estimators achieve the lowest MAE in several other operating ranges, indicating that no single information source is uniformly superior.

The refined fusion estimator was then reintegrated into EViDS together with the motion-control module for online closed-loop execution. During operation, the estimated states were supplied to the controller, while the resulting commands were applied to the CarSim vehicle model to update the vehicle response. The completed signal loop demonstrates that EViDS can support the workflow from preliminary algorithm development and real-data refinement to online closed-loop execution of environment-aware intelligent chassis algorithms.

\section{Conclusion}

This paper developed EViDS, an environment–vehicle dynamics integrated co-simulation platform for intelligent chassis development, which integrates environmental and sensor information from autonomous-driving simulation with high-fidelity vehicle dynamics. The platform was evaluated through environment–dynamics interaction tests with vehicle-dynamics consistency comparisons and intelligent chassis applications. The results demonstrate that EViDS preserves vehicle-dynamics characteristics, enables environmental information to be reliably incorporated into the vehicle-dynamics domain, and supports the development and online closed-loop validation of environment-aware intelligent chassis algorithms. Overall, EViDS provides a unified simulation basis for integrating environmental information with high-fidelity vehicle dynamics in intelligent chassis development.

\vspace{6pt}
\noindent\textbf{Acknowledgments.}
This work was supported in part by the Jiangxi Provincial Major Science and Technology Special Project under Grant No.~20262ABE010004, and in part by the National Natural Science Foundation of China under Grants No.~52472452 and No.~52325212.

%
%

\end{document}